\documentclass[conference, a4paper]{IEEEtran}
\IEEEoverridecommandlockouts
\usepackage{cite}
\usepackage{amsmath,amssymb,amsfonts}
\usepackage{algorithmic}
\usepackage{graphicx}
\usepackage{textcomp}
\usepackage{xcolor}
\graphicspath{{figs/}}
\usepackage{siunitx}
\usepackage{enumitem}
\usepackage{booktabs}
\usepackage{etoolbox}
\usepackage{multirow}
\usepackage[export]{adjustbox}
\usepackage{url}
\usepackage[caption=false,font=footnotesize]{subfig}
\def\BibTeX{{\rm B\kern-.05em{\sc i\kern-.025em b}\kern-.08em
    T\kern-.1667em\lower.7ex\hbox{E}\kern-.125emX}}

\begin{document}

\title{Training DNN Model with Secret Key for Model Protection}

\author{\IEEEauthorblockN{MaungMaung AprilPyone and Hitoshi Kiya}
\IEEEauthorblockA{Tokyo Metropolitan University, Asahigaoka, Hino-shi, Tokyo, 191--0065, Japan}
}

\maketitle

\begin{abstract}
  In this paper, we propose a model protection method by using block-wise pixel shuffling with a secret key as a preprocessing technique to input images for the first time.
The protected model is built by training with such preprocessed images.
% For inference, test images are also preprocessed with the same key.
Experiment results show that the performance of the protected model is close to that of non-protected models when the key is correct, while the accuracy is severely dropped when an incorrect key is given, and the proposed model protection is robust against not only brute-force attacks but also fine-tuning attacks, while maintaining almost the same performance accuracy as that of using a non-protected model.
\end{abstract}
% We also carried out fine-tuning attacks to simulate an attacking scenario where the adversary retrains the model by a small subset of the training dataset with the intent of adapting a forged key.


\section{Introduction}

% 1 Establish a territory: bring out the importance of the subject and/or make general statements about the subject and/or present an overview on current research on the subject.
% 2 Establish a niche: oppose an existing assumption or reveal a research gap or formulate a research question or problem or continue a tradition.
% 3 Occupy the niche: sketch the intent of the own work and/or outline important characteristics of the own work; outline important results; and give a brief outlook on the structure of the paper.
% There is no doubt that deep neural networks (DNNs) have brought major breakthoughs in computer vision, speech recognition and natural language processing.
Training a successful deep neural network (DNN) model is of great value because model training requires a huge amount of data and fast computing resources (e.g., GPU-accelerated computing). Moreover, algorithms used in training a DNN model may be patented or have restricted licenses. Considering expenses on expertise, money and time to train a DNN model, it should be regarded as a kind of intellectual property. While distributing a trained model, an illegal party may obtain the model and use it for its own service.

To protect the copyright of trained DNN models, researchers have adopted digital watermarking technology to embed watermarks into DNNs~\cite{2017-ICMR-Uchida, 2019-NCA-Le, 2019-NIPS-Fan, 2019-MIPR-Sakazawa, 2018-Arxiv-Rouhani, 2018-ACCCS-Zhang, 2018-Arxiv-Chen, 2018-USENIX-Yossi}.
These works focus on identifying the ownership of a model in question.
In reality, a stolen model can still be exploited in many different ways such as model inversion attacks~\cite{2015-CCCS-Fredrikson}, adversarial attacks~\cite{2014-ICLR-Szegedy}, internal business gains, etc.\ without being suspicious.
To the best of our knowledge, consequences of a stolen model is not considered before in model protection research except for ownership verification.
In this work, we focus on protecting a model from misuses when the model is stolen by taking an inspiration from an adversarial defense.

Recently, a key-based adversarial defense was proposed to combat adversarial examples~\cite{2020-ICIP-Maung}, that was in turn inspired by perceptual image encryption methods, which were proposed for privacy-preserving machine learning and encryption-then-compression systems~\cite{2018-ICCETW-Tanaka, 2019-Access-Warit, 2019-TIFS-Chuman, 2019-ICIP-Warit, 2019-APSIPAT-Warit, 2017-IEICE-Kurihara}.
The uniqueness of the key towards the model in the work~\cite{2020-ICIP-Maung} motivates us to adopt a key-based transformation technique for model protection.

Therefore, in this work, we propose a model protection method with a key in such a way that a stolen model cannot be used without the key for the first time.
Specifically, the proposed method preprocesses input images with a secret key and trains the model by using such preprocessed images. The preprocessing technique used in the proposed method is a low-cost block-wise pixel shuffling operation. In addition, the proposed method does not modify the network, and therefore, there is no overhead in both training and inference time. In an experiment, the performance of a protected model by the proposed method is demonstrated not only to be close to that of non-protected one when the key is correct, but also to be significantly dropped upon incorrect keys.

\section{Preliminaries}
\subsection{Model Protection}

\subsubsection{Ownership Verification}
Ownership verification is a concept to protect intellectual property of DNN models, in which digital watermarking techniques are adopted to embed watermarks into DNN models like copyright protection of media contents.
A rightful owner embeds a watermark into a model during training of the model by an embedding regularizer.
When the model is in question, the watermark is extracted with an embedding parameter from the model.
The ownership is verified by using the extracted watermark to detect the intellectual property of copyright infringement.

\subsubsection{Key-based Model Protection}
In ownership verification schemes, the performance of the protected model (i.e., fidelity) is independent of the embedded watermark.
In other words, stolen models protected with watermarking can still work well, regardless of the ownership.
The copyright of the ownership can only be claimed when the stolen model is in question.
As Fan et.\ al~\cite{2019-NIPS-Fan} pointed out, conventional ownership verification schemes are vulnerable against ambiguity attacks~\cite{1998-IEEEJSAC-Craver}.
Therefore, they proposed passport layers in the network and passports to defend ambiguity attacks.
In this work, we introduce a novel concept of model protection that is controlled by a key for the first time.
The proposed model protection does not require any additional layer in the network.
We call such a model protection method as key-based model protection.

\subsection{Related Work}
There are two forms of model protection of DNN models: white-box and black-box.
\subsubsection{White-box}
In this ownership verification scheme, the watermark is embedded to model weights by an embedding regularizer during training.
Therefore, the access to the model weights is required for extracting the watermark embedded in the model as in~\cite{2017-ICMR-Uchida, 2018-Arxiv-Chen, 2018-Arxiv-Rouhani, 2019-NIPS-Fan}.

\subsubsection{Black-box}
An inspector observes the input and output of a model in doubt to verify the ownership in black-box ownership verification schemes.
These black-box approaches exploits adversarial examples as a backdoor trigger set~\cite{2018-USENIX-Yossi, 2018-ACCCS-Zhang}, or a set of training examples is utilized in a way that a watermark pattern can be extracted from the inference of the model by the specific set of training examples\cite{2019-NIPS-Fan, 2019-MIPR-Sakazawa, 2019-NCA-Le}.
Therefore, the access to the model weights is not required to verify the ownership in the black-box approaches.

Although most of all conventional works focus on ownership identification, a state-of-the-art work~\cite{2019-NIPS-Fan} also introduced the notion of passports (i.e., a stolen model cannot be used without a correct passport).
However, the passport in~\cite{2019-NIPS-Fan} is a set of extracted features from a pre-trained model where an input is an image or a set of images, or equivalent random patterns.
In addition, the network is modified with passport layers that use passports.
Therefore, there are significant overhead costs in both training and inference time, in addition to user-unfriendly management of lengthy passports in~\cite{2019-NIPS-Fan}.

\section{Proposed Method}
\subsection{Overview}
\begin{figure}[htbp]
  \centerline{\includegraphics[width=\linewidth]{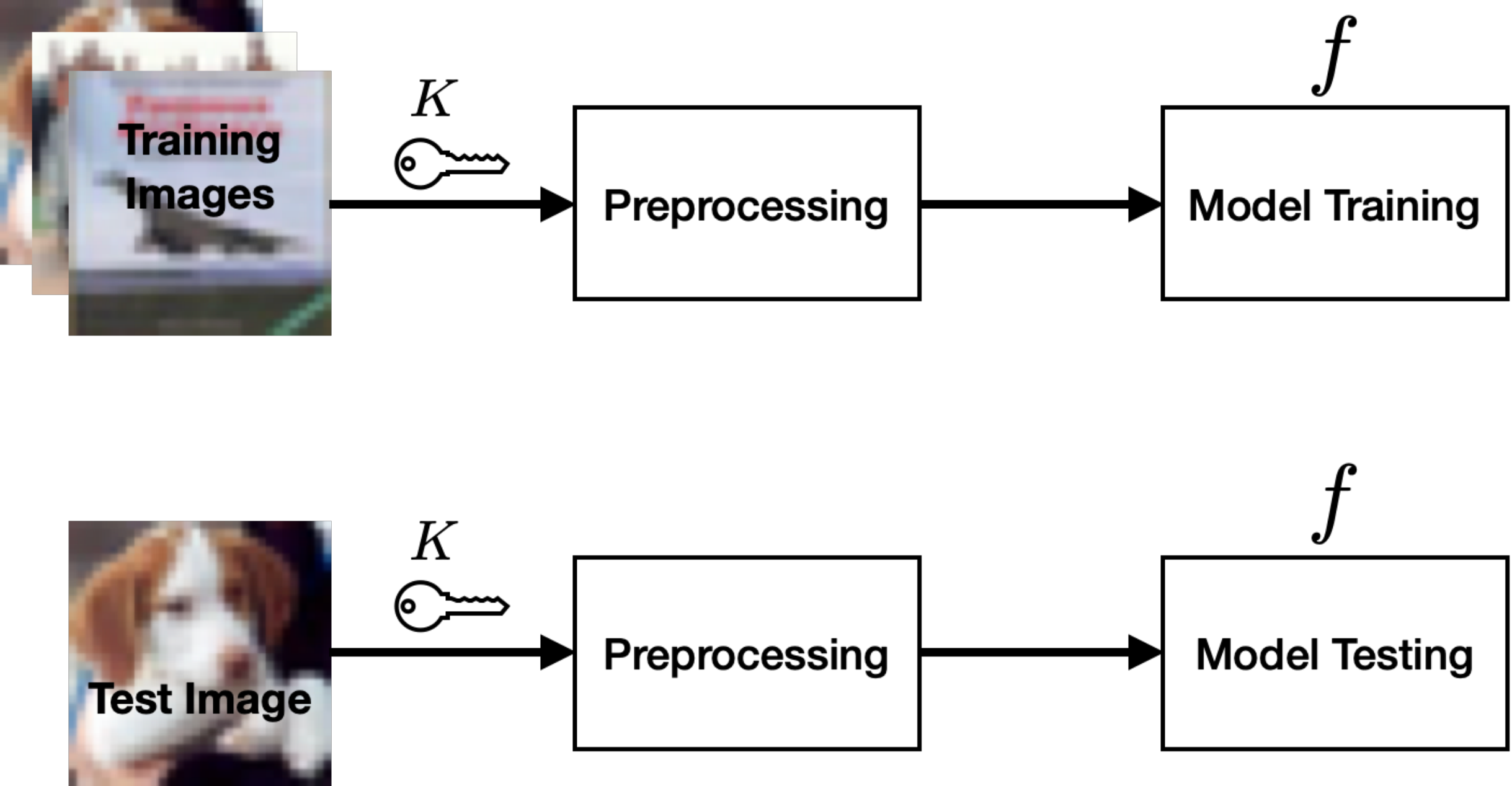}}
  \caption{Overview of proposed model protection method.\label{fig:overview}}
\end{figure}

The idea of the proposed model protection method is to preprocess input images with a secret key before training or testing a model.
An overview of the proposed method is depicted in Fig.~\ref{fig:overview}.
A model ($f$) is trained by preprocessed images with a key ($K$) for the first time.
To test the trained model, test images are also preprocessed with the same key $K$ before testing.
The next subsection details the preprocessing that is used in the proposed model protection method.
This work has been inspired by perceptual encryption methods~\cite{2018-ICCETW-Tanaka, 2019-Access-Warit, 2019-TIFS-Chuman, 2019-ICIP-Warit, 2019-APSIPAT-Warit, 2017-IEICE-Kurihara} including an adversarial defense~\cite{2020-ICIP-Maung}.

\subsection{Preprocessing\label{sec:preprocessing}}
\begin{figure*}[htbp]
  \centerline{\includegraphics[width=\linewidth]{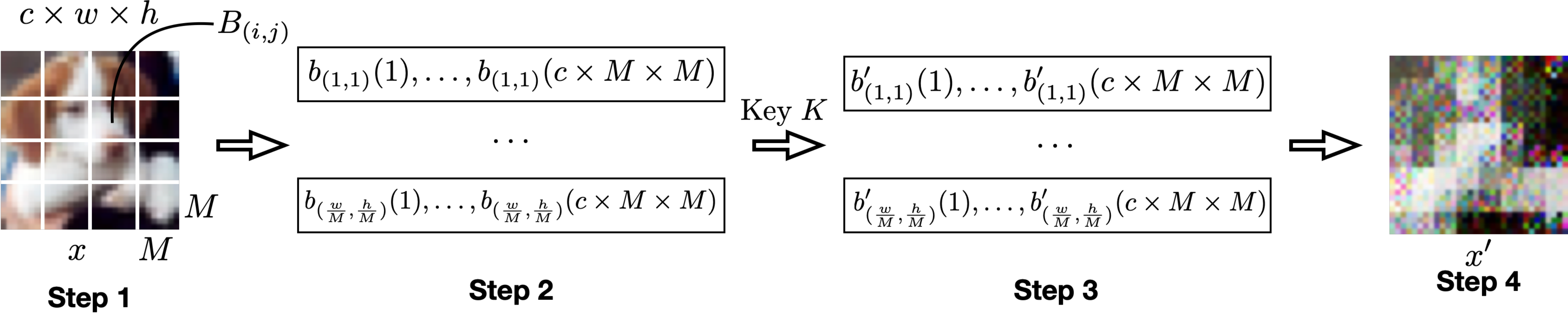}}
  \caption{Preprocessing of proposed model protection method.\label{fig:steps}}
\end{figure*}

Basically, block-wise pixel shuffling operation with a secret key is utilized as a preprocessing technique in the proposed model protection method.
The following are steps for preprocessing input images, where $c$, $w$ and $h$ denote the number of channels, width and height of an image tensor $x \in {[0, 1]}^{c \times w \times h}$ (see also Fig.~\ref{fig:steps}).

\begin{enumerate}
  \item Divide $x$ into blocks with a size of $M$ such that $\{B_{(1,1)}, \ldots, B_{(\frac{w}{M}, \frac{h}{M})}\}$.
  \item Transform each block tensor $B_{(i, j)}$ into a vector $b_{(i,j)} = [b_{(i,j)}(1), \ldots, b_{(i,j)}(c \times M \times M)]$.
  \item Generate a random permutation vector $v = [v_1, \dots, v_k, \dots, v_{k'}, \dots, v_{c \times M \times M}]$ with a key ($K$), where $v_k \neq v_{k'}$ if $k \neq k'$, and permutate every vector $b_{(i, j)}$ with $v$ as
    \begin{equation}
    b'_{(i, j)}(k) = b_{(i, j)}(v_k),
    \end{equation}
   to obtain a shuffled vector $b'_{(i, j)} = [b'_{(i,j)}(1), \dots, b'_{(i,j)}(c \times M \times M)]$
  \item Integrate the shuffled vectors to form a shuffled image tensor $x' \in {[0, 1]}^{c \times w \times h}$.
\end{enumerate}

\begin{figure}[!htbp]
\centering
\subfloat[]{\includegraphics[width=0.25\linewidth]{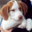}%
\label{fig:dog}}
\hfil
\subfloat[]{\includegraphics[width=0.25\linewidth]{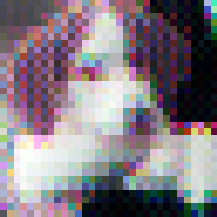}%
\label{fig:dog2}}
\hfil
\subfloat[]{\includegraphics[width=0.25\linewidth]{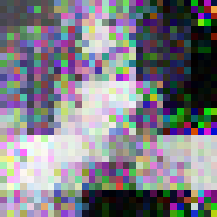}%
\label{fig:dog4}}
\hfil
\subfloat[]{\includegraphics[width=0.25\linewidth]{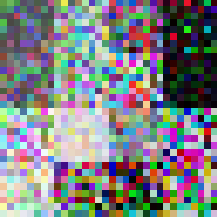}%
\label{fig:dog8}}
\caption{Example of preprocessed images in different block sizes. (a) Original image. (b) $M = 2$. (c) $M = 4$. (d) $M = 8$.\label{fig:examples}}
\end{figure}

For visualization purposes, example of preprocessed images in different block sizes are shown in Fig.~\ref{fig:examples}.

\subsection{Key Space}
Key space $\mathcal{K}$ of key $K$ used in the proposed preprocessing depends on the number of pixels in a block.
It is defined as
\begin{equation}
  \mathcal{K}(c \times M \times M) = (c \times M \times M)!.
\end{equation}
Therefore, with respect to block size $M$, the key space will vary.

\subsection{Training/Testing a Model}
A model ($f$), is trained by minibatches of images throughout the dataset iteratively.
Let a minibatch ($S$) be $\{(x^{(1)},y^{(1)}),\ldots, (x^{(m)}, y^{(m)})\}$ with $m$ samples, where $(x^{(n)}, y^{(n)})$ is a pair of an input image and a class label of $n^{\text{th}}$ sample ($n \in \{1,\ldots,m\}$).
During training, each batch is first augmented as in a normal training procedure (e.g., random cropping, random horizontal flip) and then, preprocessed with a secret key as shown in Section~\ref{sec:preprocessing} to obtain a preprocessed minibatch $S' = \{(x'^{(1)},y^{(1)}),\ldots, (x'^{(m)}, y^{(m)})\}$.
Model $f$ is trained by using all the preprocessed minibatch $S'$ of the dataset.

Similarly, a test image is also preprocessed with the same key without any augmentation.

\subsection{Requirements for Key-based Model Protection}
We consider a model protection scenario that aims to fulfill the following requirements:
\begin{enumerate}
  \item Usability: A rightful user with key $K$ can access the model without any noticeable overhead in both training and inference time, and performance degradation. The key management should be easy.
  \item Unusability: Ideally, stolen models should not be usable in any case without key $K$.
    In addition, even when the adversary retrains a stolen model with a forged key, the performance of the model should be heavily dropped.
\end{enumerate}

% \subsection{Ownership Verification}
% Verification process does not require to access the weights of the model, therefore, the proposed method is a black-box.
% The ownership is verified by the performance of the model in question on a given test dataset $D_t$ and keys.
% Assume a model $f'$ is in question, $\alpha$ is a correct key and $\beta$ is an incorrect one.
% The performance of the correct key is greater than that of the incorrect counterpart, i.e.,
% \begin{equation}
%   f'(D_t, \alpha) > f'(D_t, \beta).
% \end{equation}

\section{Experiment Results}
\subsection{Experiment Set-up}
We conducted image classification experiments on CIFAR-10 dataset~\cite{2009-Report-Krizhevsky} with a batch size of 128 and live augmentation (random cropping with padding of 4 and random horizontal flip) on a training set.
CIFAR-10 consists of 60,000 color images (dimension of $32 \times 32 \times 3$) with 10 classes (6000 images for each class) where 50,000 images are for training and 10,000 for testing.
We used deep residual networks~\cite{2016-CVPR-He} with 18 layers (ResNet18) and trained for $200$ epochs with cyclic learning rates~\cite{2017-Arxiv-Smith} and mixed precision training~\cite{2017-Arxiv-Micikevicius}.
The parameters of the stochastic gradient descent (SGD) optimizer were: momentum of $0.9$, weight decay of $0.0005$ and maximum learning rate of $0.2$.

\robustify\bfseries
\sisetup{table-parse-only,detect-weight=true,detect-inline-weight=text,round-mode=places,round-precision=2}
\begin{table*}[htbp]
\centering
\caption{Accuracy (\SI{}{\percent}) of the protected models and baseline model comparing with a conventional method\label{tab:results}}
% \resizebox{\columnwidth}{!}{%
\begin{tabular}{lSSSSSS}

  \toprule
  {Model} & {Correct} & {Incorrect} & {Plain} & {Attack 1} & {Attack 2} & {Attack 3}\\
          & {$K$} & {$K'$} & & {$\left| D' \right| = 100$} & {$\left| D' \right| = 500$} & {$\left| D' \right| = 1000$}\\
  \midrule
  {$M = 2$} & 94.70 & 25.84 & 34.39 & 18.59 & 34.44 & 46.45\\
  {$M = 4$} & 92.26 & 20.01 & 27.11 & 13.33 & 30.54 & 45.46\\
  {$M = 8$} & 86.98 & 14.98 & 15.70 & 12.57 & 33.08 & 40.22\\
  \midrule
  {Baseline} & \multicolumn{5}{c}{95.45 (Not protected)}\\
  \midrule
  {Scheme $\mathcal{V}_1$}~\cite{2019-NIPS-Fan} & 94.62 & 10 & \multicolumn{3}{l}{(Used passports instead of a key)}\\

  \bottomrule
\end{tabular}
% }
\end{table*}

\subsection{Results}
We trained three protected models by preprocessed images with a secret key in different block sizes ($M \in \{2, 4, 8\}$), and also a non-protected model (i.e., baseline model).
To verify the effectiveness of the proposed model protection method, we tested the protected models against a wrong key $K'$, plain images (without preprocessing), and fine-tuning attacks to adapt a wrong key $K'$.
Table~\ref{tab:results} summarizes simulation results.

\textbf{Usability:} When key $K$ was given, the performance accuracy was closer to the baseline accuracy.
The management of the key is straight forward and there is no noticeable overhead in both training and inference time.

\textbf{Unusability:} When the key was not correct, the accuracy was drastically decreased.
Moreover, when the protected models were tested with plain images, the performance accuracy was also significantly reduced, that implies the protected models are not usable without preprocessing input images with key $K$.

Comparing with Scheme $\mathcal{V}_1$~\cite{2019-NIPS-Fan} which is an ownership verification method with passports, the accuracy of the proposed protected model is comparable.
However, $\mathcal{V}_1$~\cite{2019-NIPS-Fan} needs overheads in both training and inference time for having passport layers in the network, and management of lengthy passports.
In contrast, the proposed model protection method does not require any overhead and has a simpler key management.

\subsection{Robustness Against Fine-tuning Attacks}
Fine-tuning (transfer learning)~\cite{2015-ICLR-Simonyan} is training a model on top of pre-trained weights.
Since fine-tuning alters weights of the model, an attacker may use fine-tuning as an attack to overwrite the protected model with the intent of forging a key.
We considered such an attacking scenario (fine-tuning attack) where the adversary has a subset of dataset $D'$ and retrains the model with a forged key ($K'$) for $30$ epochs.
We ran an experiment with different sizes of the adversary's dataset (i.e., $\left| D' \right| \in \{100, 500, 1000\}$).
Although the accuracy improved with respect to the size of the adversary's dataset, it was still way lower than the performance of the correct key as presented in Table~\ref{tab:results}.

\subsection{Future Work for Key Security}
As the key is the nucleus of the proposed model protection method, the key should be secure and hard to be estimated.
An attacker may approximate the correct key after stealing the model.
Since the attacker has stolen the model, he may try many different keys or improve a key by observing the accuracy or the loss.
We shall investigate key sensitivity of the proposed model protection method and also improve the key space of the proposed method in our future work.

% \section{Discussion}
% \subsection{Fidelity}
% Fidelity of digital watermarking in image domain is perceptual quality of the cover work.
% For model protection, fidelity is regarded as the performance of the protected model.
% As coventional model protection methods focus on ownership verification of the stolen model in question, fidelity is independent from the ownership.
% As long as the stolen model is not being questioned, it can be abused in many different ways.
% For example, the stolen model can be exploited for adversarial attacks, model inversion, internal business gains, etc.\ even before the model is suspicious for copyright infringement.
% Therefore, a scheme in a recent work~\cite{2019-NIPS-Fan} came up with the notion of passports in a way that the stolen model cannot be used without correct passports.
% However, it introduces additional overhead costs in both training and inference time, as well as lengthy management of passports.
% In contrast, the proposed model protection method uses a simple secret key and there is no significant overhead costs in both training and inference time.
% The proposed model protection method focuses on model unusability when a correct key is absent, instead of ownership verification.
% In this light, the fidelity of the protected model for the proposed method is dependent to the key.

\section{Conclusion}
We proposed a model protection method that utilizes block-wise pixel shuffling with a key to preprocess input images.
The performance accuracy of the protected model was closer to that of non-protected model when the key was correct and it dropped drastically when the incorrect key was given suggesting the protected model is not usable even when the model is stolen.
Moreover, the proposed model protection method does not introduce any overhead in both training and inference time.
The proposed method is also robust against fine-tuning attacks considering the adversary has a small subset of training dataset to adapt a new forged key.

\bibliographystyle{IEEEtran}
% \bibliography{IEEEabrv,refs}
\bibliography{IEEEabrv,../readings/references}

\end{document}